\documentclass[letterpaper, 10 pt, conference]{ieeeconf}  

\IEEEoverridecommandlockouts                              

\usepackage{graphics} 
\usepackage{epstopdf}
\usepackage{epsfig} 
\usepackage{subfigure}
\usepackage{algorithm}
\usepackage{algorithmicx}
\usepackage{algpseudocode}
\usepackage{multirow}
\usepackage{booktabs}
\usepackage{amsmath} 

\title{\LARGE \bf
A Unified Framework for Mutual Improvement of SLAM and Semantic Segmentation
}

\author{Kai Wang$^{1}$  Yimin Lin$^{1}$  Luowei Wang$^{1}$  Liming Han$^{1}$
 Minjie Hua$^{1}$ Xiang Wang$^{1}$ Shiguo Lian$^{1}$  Bill Huang$^{1}$
\thanks{$^{1}$ All the authors are with CloudMinds Technologies Inc., Beijing 100102, China.
        {\tt\small kai.wang,anson.lin,luowei.wang,
        liming.han,michael.hua,xiang.wang,scott.lian,
        bill@cloudminds.com}}%
}

\begin{document}

\maketitle
\thispagestyle{empty}
\pagestyle{empty}

\begin{abstract}
This paper presents a novel framework for simultaneously implementing localization and segmentation, which are two of the most important vision-based tasks for robotics. While the goals and techniques used for them were considered to be different previously, we show that by making use of the intermediate results of the two modules, their performance can be enhanced at the same time. Our framework is able to handle both the instantaneous motion and long-term changes of instances in localization with the help of the segmentation result, which also benefits from the refined 3D pose information. We conduct experiments on various datasets, and prove that our framework works effectively on improving the precision and robustness of the two tasks and outperforms existing localization and segmentation algorithms.
\end{abstract}

\section{INTRODUCTION}\label{sec:Intro}

Localization and Segmentation are two of the most fundamental tasks for robotic movement and sensing. The former computes the robot's current position and orientation, and the latter helps to perceive the distribution and precise boundaries of the objects of interest within the robot's field of view. These two techniques are essential in many robotic applications including autonomous driving, Unmanned Aerial Vehicles (UAV), robot patrolling and logistics, etc.

For the localization task, visual Simultaneous Localization and Mapping (vSLAM) is one of the most promising methods due to its relatively low hardware and computational cost characteristics in recent years. It utilizes image sequences with some auxiliary sensor data such as depth map, Inertial Measurement Unit (IMU) data, etc., to create the map
of the environment and return the current location information at the same time. A big challenge in vSLAM is that the environment in which the robot locates is usually changeable. On one hand, instantaneous movement of some objects during mapping will affect the precision of the map due to the inconsistency of the moving trend in the scene~\cite{DynaSLAM}. On the other hand, the map created will no longer be consistent with the environment once some objects have moved after mapping completes. As a result, subsequent localization based on this map will not be accurate.

For the segmentation task, 2D image-based semantic segmentation using deep neural network has proved to be effective in most cases and has been widely used in many systems~\cite{segmensurvey}. It is able to output the exact boundaries of a series of segmented regions and their classes. Anyway, unprecise manual labeling and lack of similar training data usually lead to inaccurate segmentation results for these deep learning methods.

Previously, these two tasks were generally regarded as two independent tasks whose results were rarely utilized by each other. In this paper, we propose a novel framework for simultaneously improving the vSLAM as well as semantic segmentation precisions. The segmentation and vSLAM are performed in an interweaved method and the results are used to refine each other's. Specifically, the computed pose information of the previous and current frames are utilized to refine the segmentation of the latter one, in which all the potentially moveable objects are then identified and sent to the vSLAM module for further computation of the tracking and mapping of the corresponding frame. This scheme repeats through the whole process and both the vSLAM and segmentation precisions of this sequence are therefore enhanced. Furthermore, the map created becomes more robust to changes of the scene and the localization in the same environment afterwards will benefit from it and become more precise. This framework is tested on different datasets and proves to be more effective over existing works on both the vSLAM and segmentation tasks.


The contributions of this paper include:

\begin{itemize}
  \item A unified framework of enhancing the vSLAM and segmentation tasks mutually.
  \item A novel approach for enhancing both the mapping and localization precisions in vSLAM by
  identifying and processing both the moving and potentially moveable objects respectively.
  \item An effective refinement scheme for image segmentation by making use of 3D pose information.
\end{itemize}

The rest of the paper is organized as follows: Section~\ref{sec:review} reviews the
related works on vSLAM and segmentation. Section~\ref{sec:algo} introduces the proposed
framework and workflow in detail, and experimental results are shown and
discussed in Section~\ref{sec:res}. Section~\ref{sec:conc} gives the conclusion.




\section{RELATED WORK}\label{sec:review}

\begin{figure*}[t]
\centering
\includegraphics[scale=0.5]{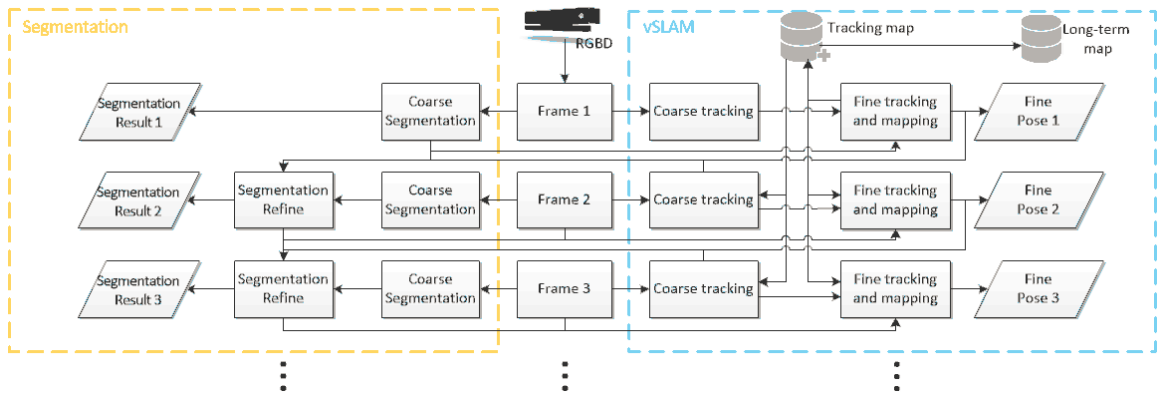}
\caption{The overall workflow of the proposed framework, which contains a segmentation module and a vSLAM module. For each input frame, a coarse pose and segmentation are first calculated. The two results are then used to estimate a fine pose and update a tracking map. A long-term map is also maintained for the further visit of the same area. At the same time, the segmentation results can also be refined by using that of the previous frame and the poses estimated in the two frames. The refinement of the vSLAM and segmentation results is implemented within a single iteration for each frame.}
\label{fig:workflow}
\end{figure*}

\subsection{vSLAM for Dynamic Scenes}\label{sec:review:slam}

vSLAM is used to estimate the camera location and 3D map of the scene through a set of feature correspondences extracted from a series of images~\cite{SLAMsurvey}. Various works on vSLAM have been proposed in recent years, from the seminal work PTAM~\cite{PTAM} to the popular ORB-SLAM2~\cite{ORBSLAM}. Most of these approaches assume that the observed
scenes are relatively static, and pose estimation might drift or even be lost as there are not features to be matched consistently in the case of scenes with dynamic objects.

There have been works proposed to handle dynamic environments~\cite{dynamicvslamsurvey}.
For example,~\cite{opticalflow} computed the likelihood of a moving object based on a motion metric computed from optical flow and then segment the moving objects.~\cite{RS15} further extended it to handle stereo image sequences. Recently, researchers have shifted their focus to using deep neural network to do the segmentation to remove outliers for accurate pose estimation. For example, Mask-SLAM~\cite{MASKSLAM} excludes feature points detected in the sky area or on cars using the segmentation mask trained by DeepLab v2~\cite{DeepLabv2}.
The work~\cite{DynaSLAM} proposed to combine multi-view geometry models and deep-learning-based algorithms for detecting dynamic objects and removed them from the frames. In~\cite{Robust}, the depth map, sparse scene flow and semantic cues are combined to classify scene as either static background, movable and moving objects. While these methods have proved that excluding feature points in certain masked area makes the estimation of camera motion more stable, they rely heavily on the exact segmentation of the moveable objects and are prone to be inaccurate when its precision is limited. The idea in~\cite{ZW18} which identified dynamic objects to enhance the vSLAM precision and further provided a refined dataset for training the object detection network is similar to our work, except that the extraction of objects with object detection in the first step is less accurate, and the second step remains an offline scheme.


\subsection{Image and Video Segmentation}\label{sec:review:seg}

 The pioneering work~\cite{FCN} on deep neural network based image segmentation explored the use of Convolutional Neural Network (CNN) to segment the images, through adapting classifiers for dense prediction by replacing the last fully-connected layer with deconvolution layers. Later on,~\cite{SegNet} made use of the encoder-decoder architecture and reused the pooling indices from the encoder to decrease parameters. DeepLabv3~\cite{DeepLabv3} augments the Atrous Spatial Pyramid Pooling (ASPP) module in~\cite{DeepLabv2} with image-level feature to capture longer range information as in~\cite{lowtensor}, and DeepLabv3+~\cite{DeepLabv3+} further extends it to include an effective decoder module to refine the segmentation results along object boundaries. Pyramid Scene Parsing Network (PSPNet)~\cite{PSPNet} implements spatial pooling at several grid scales and demonstrates satisfactory performance.

Furthermore, algorithms have been proposed to achieve instance-level segmentation.
The prior work~\cite{XXX0} task uses R-CNN~\cite{XXX1} to classify region proposals, which are then refined by category-specific coarse mask predictions.
MNC~\cite{XXX4} proposed a cascaded structure, which consists of three networks used for differentiating instances, estimating masks, and categorizing objects respectively. FCIS~\cite{XXX5} performs object segmentation and detection sub-tasks jointly and exploits the strong correlation between the two sub-tasks with shared score maps. Mask R-CNN~\cite{XXX6} extends Faster R-CNN~\cite{XXX7} by adding a branch for predicting an object mask in parallel with the existing branch for bounding box recognition.

There are also some works proposed for video sequence-based segmentation. For example,~\cite{DST-FCN} made use of the spatial-temporal information of consecutive frames by introducing 3D-Conv~\cite{3D-Conv} and Conv-LSTM~\cite{Conv-LSTM} modules, so as to enhance the precision of video segmentation. Since the 3D spatial information of adjacent frames was not utilized, they may still fail to predict precise boundary information.




\section{Framework}\label{sec:algo}

\subsection{Overall Workflow}

The general workflow of the proposed framework is shown in Fig.~\ref{fig:workflow}. This framework takes the RGB image sequences as well as the depth map sequences as input. It includes two major modules: the vSLAM module and the segmentation module. For each input frame, the vSLAM module will output the pose information of the camera w.r.t. the world and update the map of the environment for long-term use, and the segmentation module will produce an image segmentation result with the semantic information of each pixel.

Specifically, the initial input frame will be first segmented, and potentially dynamic objects are identified. At the same time, a coarse pose is computed in the vSLAM module. The results will then be sent to the vSLAM module to build the map. Next, when a new frame comes, a coarse vSLAM and segmentation will be performed first, and the coarse pose together with the pose and segmentation result of the last frame will be sent to the segmentation module to refine the coarse result. After the final segmentation result of this frame is computed, it will be sent to the vSLAM module to proceed fine tracking and mapping, after which the precise map and location information will be obtained.

Next, the detailed vSLAM and segmentation modules will be introduced.

\subsection{Initial Segmentation}

For each input RGB frame, we used the FCIS~\cite{XXX5} algorithm which proved to be effective on various datasets to perform an initial segmentation. We trained the network on MS COCO~\cite{MSCOCO} dataset which contains 80 classes for both indoor and outdoor objects. For an input RGB image, FCIS is able to compute the bounding box for each object. If the pixel value in the bounding box is larger than a threshold, it is regarded as part of the object; otherwise, it will be marked as the background. We repeat this operation for all the bounding boxes to get the mask for the whole image.


After the segmentation, we identified the moveable objects from all the instances in the result, according to a predefined shortlist in which only objects that are likely to move or be moved
(such as person, cars, cup, chair, etc.) among all the 80 classes are selected. The result is in the form of a mask image with the region and instance ID of each segmented instance encoded, and will be sent to the vSLAM module to proceed the tracking and mapping computation.

\subsection{vSLAM based on Segmentation Result}

We use the ORB-SLAM2 algorithm~\cite{ORBSLAM} which has shown satisfactory performance in many scenarios. To ensure the stability, we used the RGB-D version of ORB-SLAM2 which takes both RGB image and depth map as input.

Each time a new frame comes, we first implement a coarse tracking to get an initial guess of the pose of the current frame. Specifically, we first extract the ORB feature points and align them with the depth map to get the 3D coordinates $(P_x, P_y, P_z)$ of each point $P$, and get the coarse rotation $R_c$ and translation $T_c$ by minimizing the reprojection error as what the original ORB-SLAM2 did.

The extracted feature points are then classified into a background set $A$ and other different sets $\{B_i| i=1...n\}$ according to their positions in different segmented areas. If a point $P$ lies in the background area, it belongs to set $A$; otherwise it falls into set $B_i$ which corresponds to the area of segmented instance $i$. The motion states of the classified point sets will then be judged according to the coarse rotation $R_c$ and translation $T_c$. Specifically, we project the points in the tracking map onto the current frame, and for each point $P_i$ in the frame, a best matching point $P_{match}$ is found. If the Euclidean distance between $P_i$ and  $P_{match}$ is less than a predefined threshold, then $P_i$ is regarded as static. For the set $B_i$ that $P_i$ belongs to, if the percentage of moving points is less than a threshold, then the instance that set $B_i$ corresponds to is regarded as a static object in the current frame, otherwise, it is deemed moving.

\begin{figure} [htbp]
  \centering
  \subfigure[]{
    \centering
    \label{fig:classify:a} 
    \begin{minipage}[b]{0.2\textwidth}
      \centering
      \includegraphics[scale=0.4]{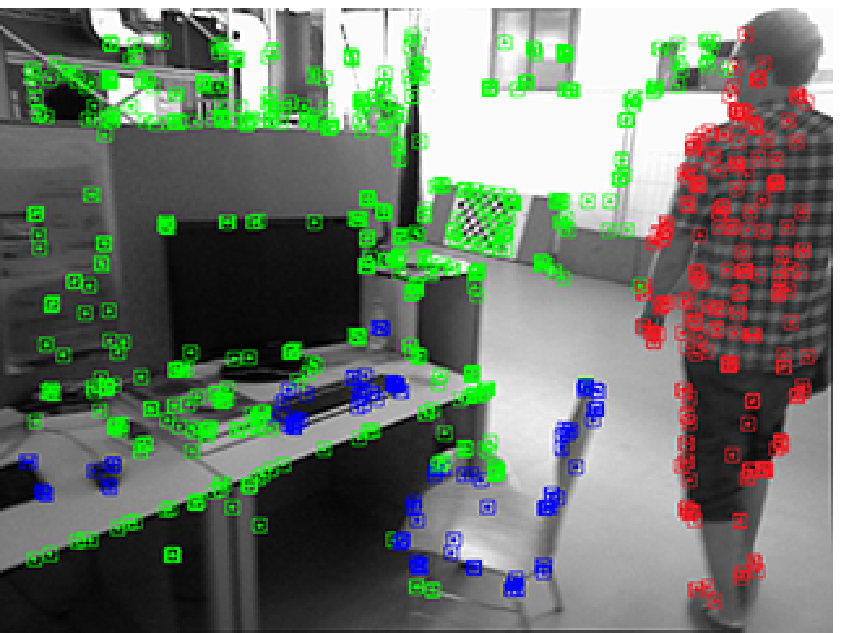}
    \end{minipage}}
  \subfigure[]{
    \centering
    \label{fig:classify:b}
    \begin{minipage}[b]{0.2\textwidth}
      \centering
      \includegraphics[scale=0.4]{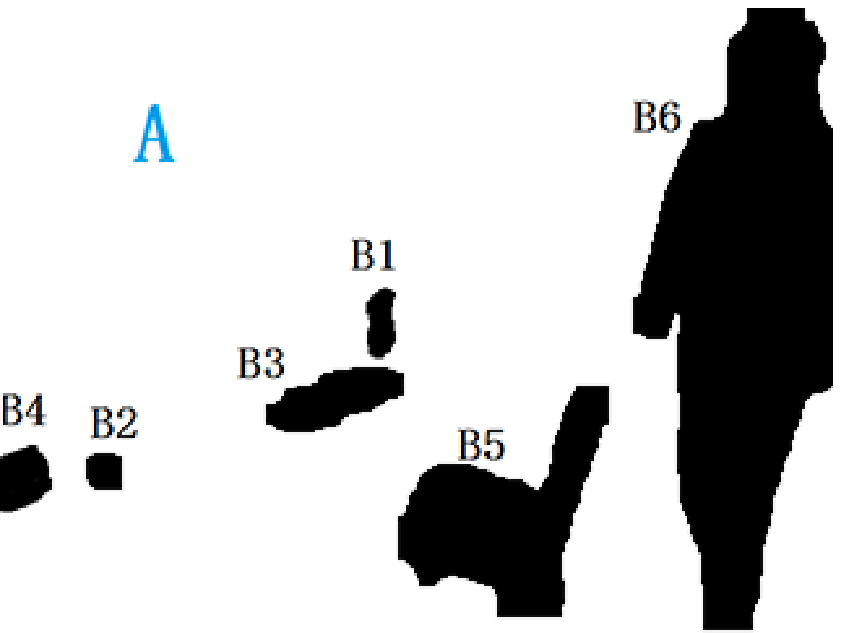}
    \end{minipage}}
  \caption{Illustration of feature points and segmented area classifications.
  (a) The detected feature points are classified into background (in green), moving (in red) and moveable (in blue) points;
  (b) The segmentation result with regions classified into background (A) and moving or moveable (B1-B6) . }
  \label{fig:classify}
\end{figure}

An example of the segmented regions and classified features points is shown in Fig.~\ref{fig:classify}.

Next, 2D-3D matching between the points in the background set $A$ and the sets $\{B_s\}$ that are considered as static and also in the tracking map is implemented by minimizing the reprojection error, and fine rotation $R_f$ and translation $T_f$ can thus be obtained. After the fine pose has been obtained, it will be sent to the segmentation module for the refinement of the initial segmentation result.

There are two types of maps created and maintained in the vSLAM module: tracking map and long-term map.

The tracking map is used to compute the trajectory of the camera during the tracking process.
The new point $P_m$ in the tracking map is computed by projecting each point $P_c$ the background point set $A$ and
moving point set $B_s$ of the new key frame onto the tracking map through $P_m=R_f P_c + T_f$. If there are already matching points, then no more update of the map is required; otherwise, the newly projected 3D points will be added into the tracking map. The use of only points of the static objects will help the preservation of the information used for computing the camera pose in the current scene, and thus improves the tracking stability and trajectory precision.

The long-term map is designed for long-term use. It only needs to be created at the first time when a robot
navigates in a new area, and can be reused later on to avoid duplicated mapping computation when the same region is visited . Therefore, only the points whose positions will probably remain fixed over
time should be included in this map to provide stable environment information. To do that, each time the tracking
map is updated, we remove the points that belong to set $B_s$ in the tracking map and have the potential
to move in future, and add the rest points (i.e. the points in set $A$) into the long-term map.

\subsection{Refinement of Segmentation Result}

After we get the coarse pose $R_c$, $T_c$ of the current frame, and the fine pose $R_f$, $T_f$
as well as the segmentation result of the previous frame, we can use them to update the segmentation
result in the current frame.

First, we project each 2D point $(p_u, p_v)$ of the segmented regions in the last frame which has been refined and assumed to be accurate to $(p_u^\prime, p_v^\prime)$ in the current frame  according to the following equations:

\begin{eqnarray}\label{eq:proj}
  {P_z} &=& D({p_u},{p_v})/DF ,\\
  {P_x} &=& ({p_u} - {c_x})*{P_z}/{f_x} ,\\
  {P_y} &=& ({p_v} - {c_y})*{P_z}/{f_y} ,\\
  \left[ \begin{array}{l}
p_u^\prime\\
p_v^\prime
\end{array} \right] &=&
\left[ {\begin{array}{*{20}{c}}
{{f_x}}&{{0}}&{{c_x}}\\
{{0}}&{{f_y}}&{{c_y}}\\
{{0}}&{{0}}&{{1}}
\end{array}} \right]
\left[ {R|T} \right]
\left[ \begin{array}{l}
P_x\\
P_y\\
P_z\\
1
\end{array} \right]
/s .
\end{eqnarray}

In the above equations, $f_x$, $f_y$ and $(c_x, c_y)$ are the focal lengths and principal point of the camera respectively. $D({p_u},{p_v})$ is the depth value of $(p_u, p_v)$ and $DF$ is the depth factor of the depth map. $R=R_c^{-1}R_f$ and $T=T_f-T_x$ represent the relative rotation and translation w.r.t. the last frame. $s$ is the scale factor of the image.

Next, we try to refine the initially segmented image with each projected region $Re_p$. The workflow is listed
in Algorithm~\ref{alg:reg}. We first try to find a matching region for $Re_p$ in the current frame, by measuring
the similarity $S_{cp}$ between each region $Re_c$ in the roughly segmentation result and $Re_p$ using:

\begin{equation}\label{eq:sim}
\begin{aligned}
   S_{cp} &= w1*Dist(Re_c, Re_p) \\
          & + w2* \sqrt{\frac{Area((Re_c-Re_p)\cup (Re_p-Re_c))}{Area(Re_c)+Area(Re_p)}} ,
\end{aligned}
\end{equation}

where $Dist(Re_c, Re_p)$ refers to the Euclidean distance between the barycenters of $Re_c$ and $Re_p$. The function $Area(\cdot)$ computes the total number of pixels inside a region. The first item measures the positional distances of the two regions and the second one is used to compute their shape difference. $w1$ and $w2$ are the weights for the two items respectively. The region $Re_c$ with the smallest $S_{cp}$ value which is smaller than a predefined threshold will be selected as the matching region for $Re_p$. We then compare the ratios of their intersection area to the two regions, and the region with larger ratio is considered as a reliable one and preserved as the finally segmented region.

\begin{algorithm}[htbp]
\caption{Workflow for segmented regions' update.}
\label{alg:reg}
\begin{algorithmic}[1]
\For {a projected region $Re_p$}
    \State find the matching region $Re_c$ for $Re_p$ with~(\ref{eq:sim})
    \If {found}
        \State  compute $In_{cp}=Re_c \cap Re_c$
        \State  compute $Ra_c=In_{cp} / Re_c$
        \State  compute $Ra_p=In_{cp} / Re_p$
        \If {$Ra_c<Ra_p$}
            \State replace $Ra_c$ with $Ra_p$
        \Else
            \State //do nothing.
        \EndIf
    \Else
        \If {$\#regions^{t-1} > \#regions^t$}
        \State add $Ra_p$ to segmentation result
        \Else
            \State //do nothing.
        \EndIf
    \EndIf
\EndFor
\end{algorithmic}
\end{algorithm}

If no matching region is found for $Re_p$, there is a high possibility that the segmentation algorithm failed to recognize an instance that was supposed to be segmented when the number of segmented instances in the current frame
is less than that of the previous one. In that case, we will update the segmentation result by adding $Re_p$ to it.
If the numbers of segmented instances are same, then we simply skip the current $Re_p$ and repeat the same process for the next $Re_p$.

It should be mentioned that this strategy is based on the assumption that there is no drastic changes between two adjacent frames. In some extreme circumstances, for example, if the frequency of the camera is not high enough to ensure the fast-moving objects be well captured, the algorithm may fail on judging the region correspondence and lead to fake results. This may be alleviated by introducing frame interpolation into the computation, although this case is rarely seen in real applications.

\section{RESULTS and DISCUSSIONS}\label{sec:res}

We test our framework on different datasets with ground truths available, and compare with other state-of-the-art works on vSLAM and image segmentation. We run each sequence ten times as in~\cite{DynaSLAM} to compensate for the non-deterministic nature of dynamic scenes. All tests were implemented on a workstation with Intel i7 6700K CPU, with 32 \emph{GB} RAM and Nvidia GTX1070 GPU.

\subsection{Test Results on TUM Dataset}

We first test the performance of the vSLAM module of our framework on TUM dataset~\cite{TUM} in which 39 RGB-D sequences are collected. Each sequence contains both $640 \times 480$ 8-bit RGB images and $640 \times 480$ 16-bit depth images, with the ground truth of the camera trajectory provided. Specifically, we select 6 sequences which contain 'walking' and 'sitting' from the 'fr3' subset. The images were taken in the 'desk' scene, in which two persons are either walking or sitting, and thus are suitable for testing the efficiency of our algorithm under scenes with dynamic objects.

We compared our algorithm with the original ORB-SLAM2~\cite{ORBSLAM} and DynaSLAM~\cite{DynaSLAM} in terms of Absolute Trajectory Error (ATE)~\cite{TUM} which represents the tracking precision by taking the ground truth as reference, and the results are shown in Table~\ref{table:ate}.

\begin{table*}[htbp]
\caption{Comparisons of ATE[m] of our vSLAM module against the original ORB-SLAM2~\cite{ORBSLAM} and DynaSLAM~\cite{DynaSLAM}.}
\begin{center}
\begin{tabular}{|c|c|c|c|c|c|}
\hline
\multicolumn{1}{|c|}{\multirow{2}{*}{Sequence}} &
\multicolumn{1}{|c|}{\multirow{2}{*}{ORB-SLAM2}}&
\multicolumn{1}{|c|}{\multirow{2}{*}{DynaSLAM}}&
\multicolumn{3}{|c|}{Our vSLAM module} \\
\cline{4-6}
\multicolumn{1}{|c}{} & \multicolumn{1}{|c}{} & \multicolumn{1}{|c|}{}  & median & min & max\\
\hline Walking\_halfsphere&0.351&0.025&\textbf{0.019}&0.010&0.028\\
\hline Walking\_static&0.090&0.006&\textbf{0.005}&0.0005&0.008\\
\hline Walking\_rpy&0.662&0.035&\textbf{0.032}&0.002&0.036\\
\hline Walking\_xyz&0.459&0.015&\textbf{0.014}&0.001&0.029\\
\hline Sitting\_halfsphere&0.020&\textbf{0.017}&0.021&0.002&0.031\\
\hline Sitting\_xyz&\textbf{0.009}&0.015&\textbf{0.009}&0.001&0.022\\
\hline
\end{tabular}
\label{table:ate} 
\end{center}
\end{table*}

It can be seen from Table~\ref{table:ate} that the improvement of the performance of our algorithm on the 'walking' datasets is obvious. In these datasets, ORB-SLAM2 created a lot matches of dynamic feature points due to the movement of the two persons. This enlarges the pose error during optimization. Similar to DynaSLAM, we segmented and discarded the moving objects which contribute to the dynamic points and therefore reach higher precisions. The reason why our algorithm outperforms DynaSLAM is because we refined the segmentation results using 3D pose information and obtained more accurate segmentation regions and boundaries. The enhancement of segmentation precision makes the removal of dynamic points more accurate and thus reduces the pose error. For the 'sitting' datasets, the improvement of our algorithm is not quite obvious, as there are limited dynamic objects in that scene, which do not affect the feature points matching too much.

\begin{figure} [htbp]
  \centering
  \subfigure[]{
    \centering
    \label{fig:tum:a} 
    \begin{minipage}[b]{0.2\textwidth}
      \centering
      \includegraphics[scale=0.4]{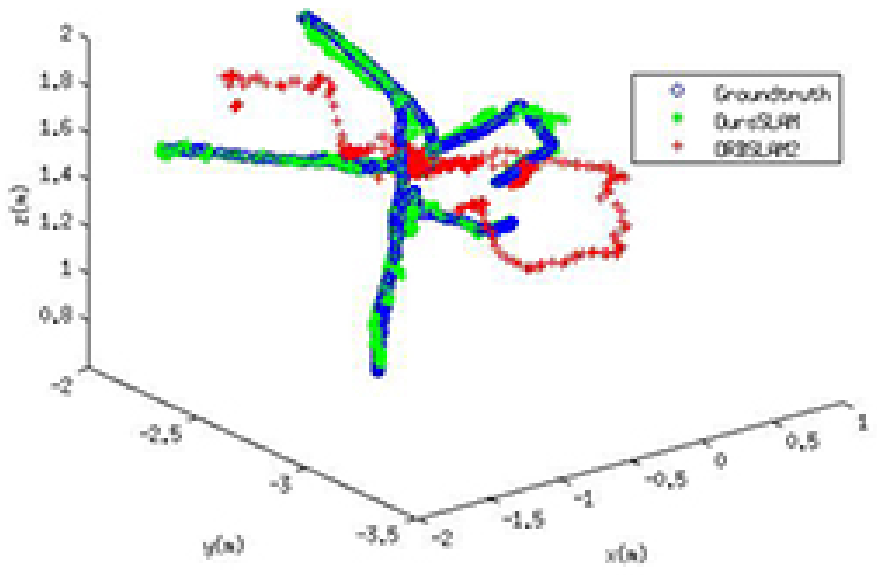}
    \end{minipage}}
  \subfigure[]{
    \centering
    \label{fig:tum:b}
    \begin{minipage}[b]{0.2\textwidth}
      \centering
      \includegraphics[scale=0.4]{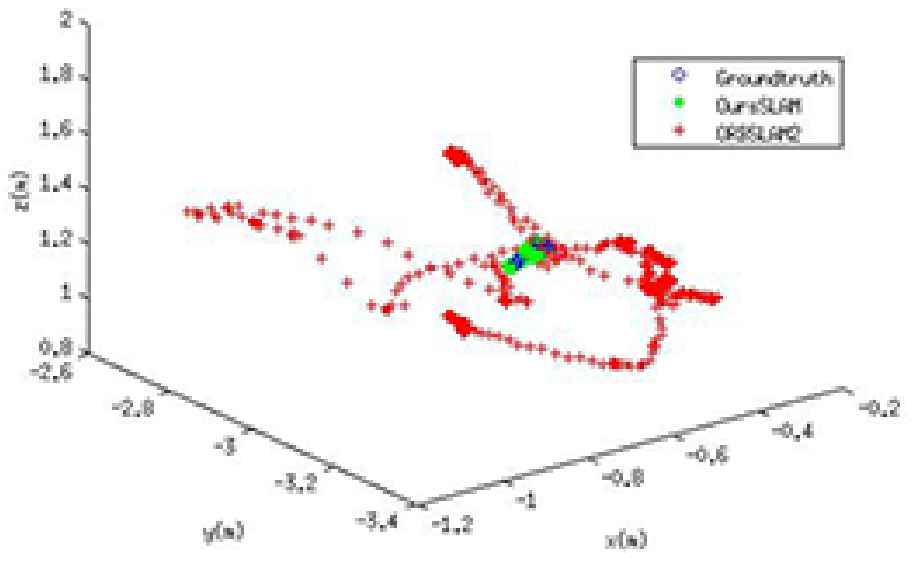}
    \end{minipage}}
  \subfigure[]{
    \centering
    \label{fig:tum:c}
    \begin{minipage}[b]{0.2\textwidth}
      \centering
      \includegraphics[scale=0.4]{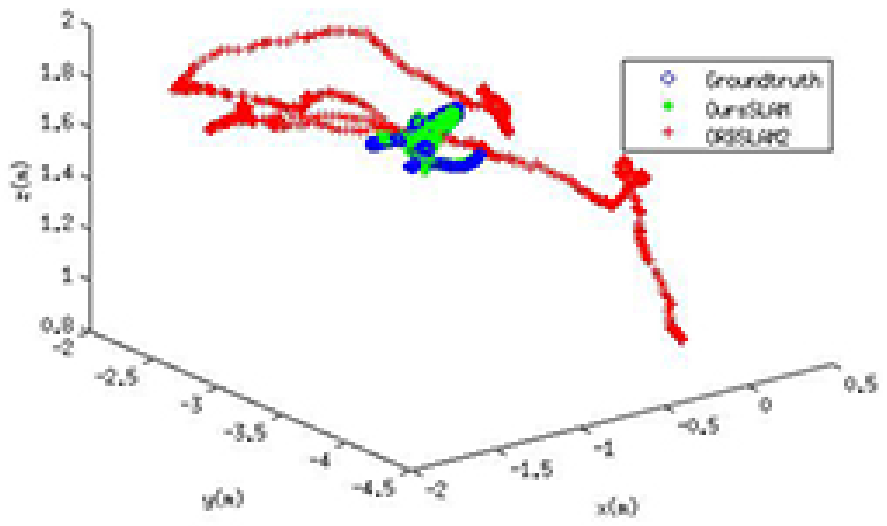}
    \end{minipage}}
  \subfigure[]{
    \centering
    \label{fig:tum:d}
    \begin{minipage}[b]{0.2\textwidth}
      \centering
      \includegraphics[scale=0.4]{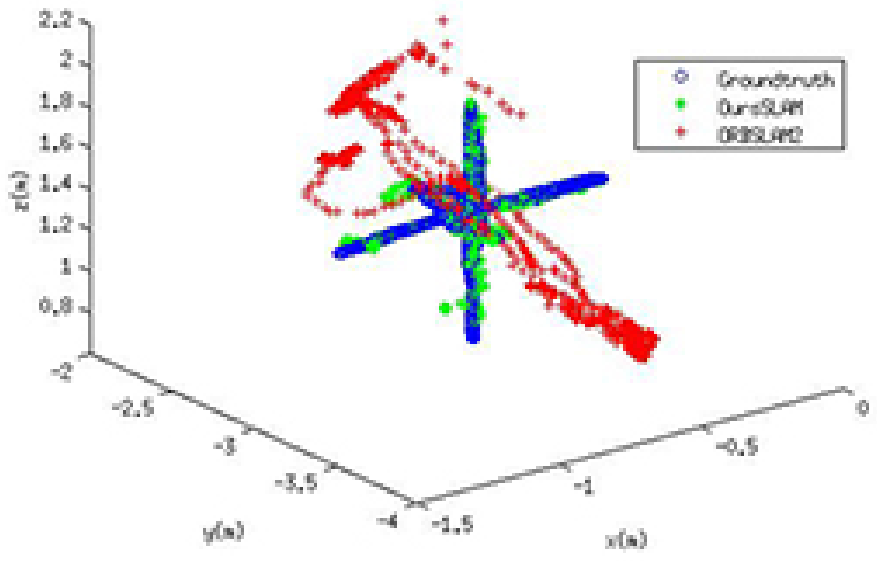}
    \end{minipage}}
  \caption{Comparison of output trajectories of our vSLAM module(in green), ORB-SLAM2~\cite{ORBSLAM}(in red) and ground truth(in blue) of the (a)'walking\_halfsphere', (b)'walking\_static', (c)'walking\_rpy' and (d)'walking\_xyz' of the TUM dataset~\cite{TUM} respectively.}
  \label{fig:tum}
\end{figure}

We also visualized the trajectory that our algorithm outputs with those of ORB-SLAM2 and ground truth in Fig.~\ref{fig:tum} with green, red and blue respectively. It can be seen that our result exhibits much higher similarity to ground truth than ORB-SLAM2 does.

The average time for the coarse tracking is 6 \emph{ms}, and the fine tracking and mapping takes 22 \emph{ms}.

\subsection{Test Results on ScanNet Dataset}

As the ground truth for segmentation is not available in TUM dataset, we used the ScanNet dataset~\cite{Scannet} to evaluate the performance of our segmentation module. ScanNet contains 1500 RGBD sequences taken in indoor environment, and has totally 2.5 million images available. The resolutions of RGB images and depth maps are $1296 \times 968$ and $640 \times 480$ respectively. With the provided extrinsic parameters, each depth map can be mapped to the RGB image. Ground truths of the segmentation is available for every RGB image. As the image sets in ScanNet has 550 object classes, we manually map each class to the MS COCO 80 classes according to its name or general type.

For all the images, we compute the mean Average Precision (mAP) and mean Intersection over Union (mIoU) for the results generated using our segmentation module and the original FCIS~\cite{XXX5} algorithm. The results are shown in Table~\ref{table:segscan}.

\begin{table}[htbp]
\caption{Comparison of FCIS~\cite{XXX5} and our segmentation module on ScanNet dataset.}
\begin{center}
\begin{tabular}{|c|c|c|}
\hline &FCIS&Our segmentation module\\
\hline mAP&0.6314&\textbf{0.6504}\\
\hline mIoU&0.5620&\textbf{0.5751}\\
\hline
\end{tabular}
\label{table:segscan} 
\end{center}
\end{table}

It can be seen that the segmentation precision of our module has been improved comparing to that of FCIS~\cite{XXX5}. This proves that the use of 3D pose information for the refinement of segmented areas works well as expected.

\begin{figure} [htbp]
  \centering
  \subfigure[]{
    \centering
    \label{fig:scannet:a} 
    \begin{minipage}[b]{0.15\textwidth}
      \centering
      \includegraphics[scale=0.26]{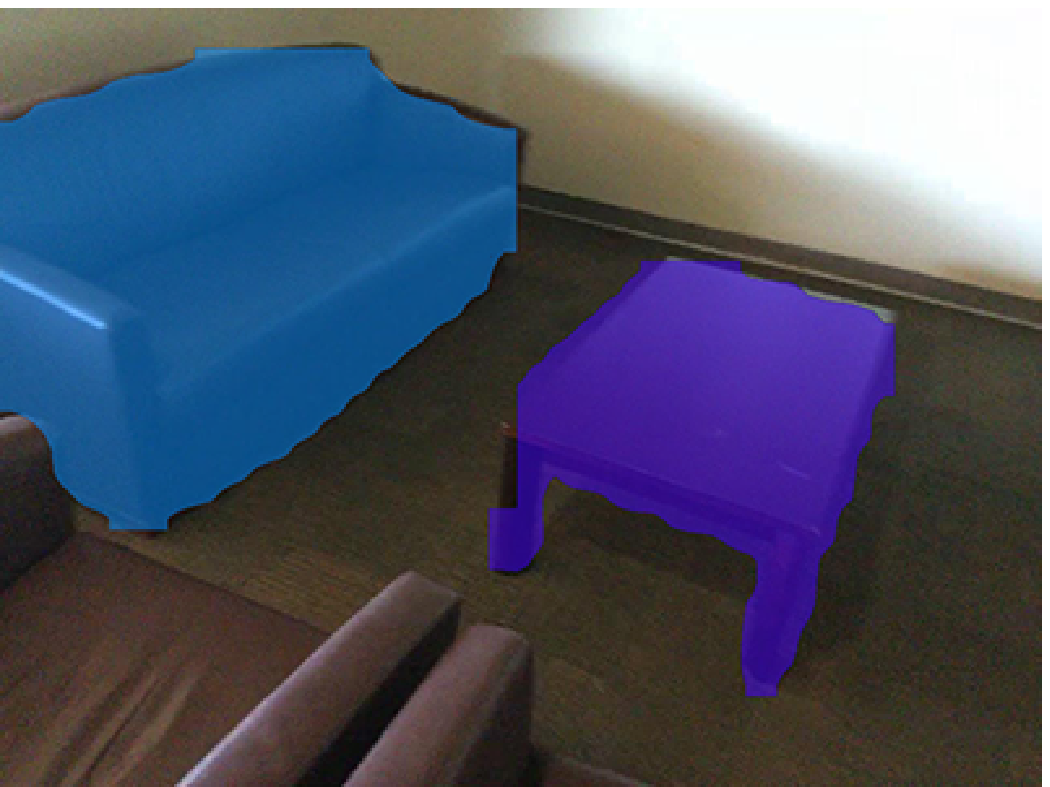}
    \end{minipage}}
  \subfigure[]{
    \centering
    \label{fig:scannet:b}
    \begin{minipage}[b]{0.15\textwidth}
      \centering
      \includegraphics[scale=0.26]{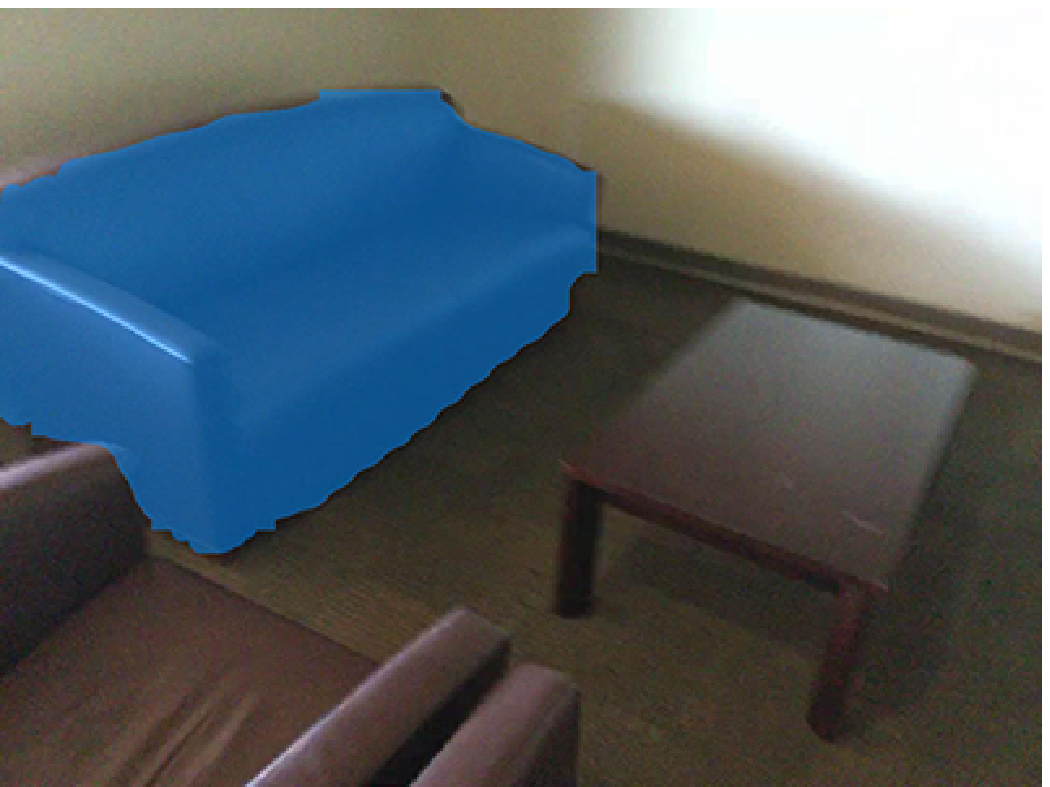}
    \end{minipage}}
  \subfigure[]{
    \centering
    \label{fig:scannet:c}
    \begin{minipage}[b]{0.15\textwidth}
      \centering
      \includegraphics[scale=0.26]{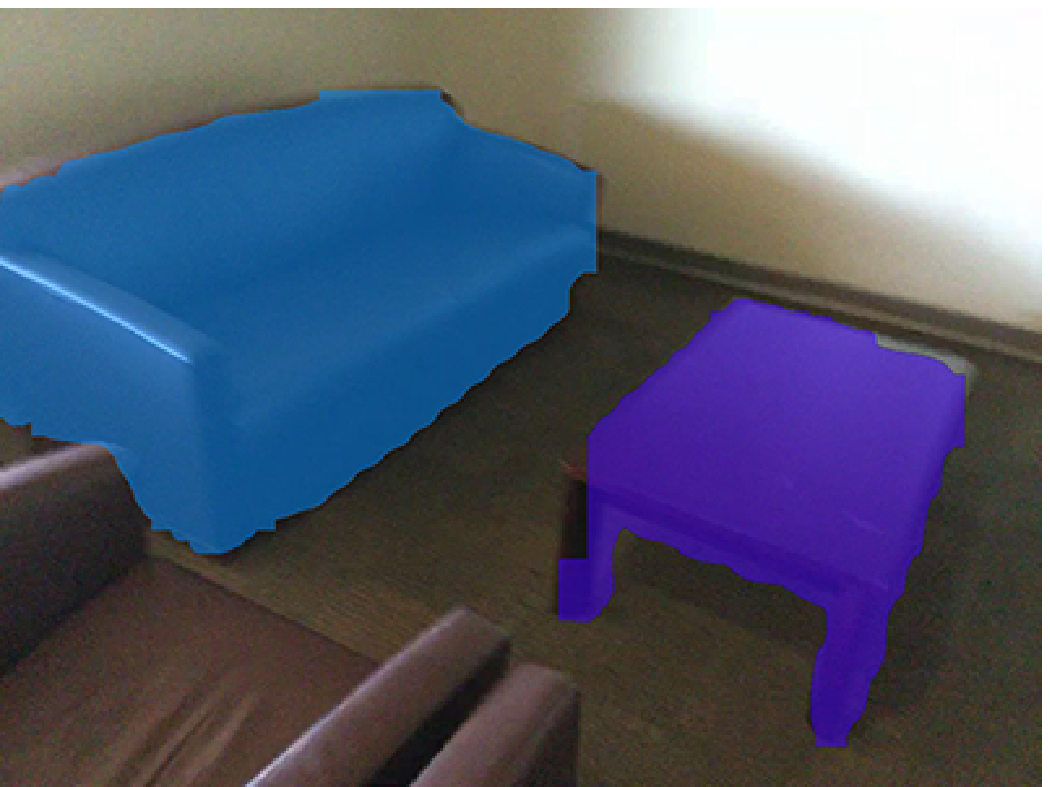}
    \end{minipage}}
  \subfigure[]{
    \centering
    \label{fig:scannet:d}
    \begin{minipage}[b]{0.15\textwidth}
      \centering
      \includegraphics[scale=0.26]{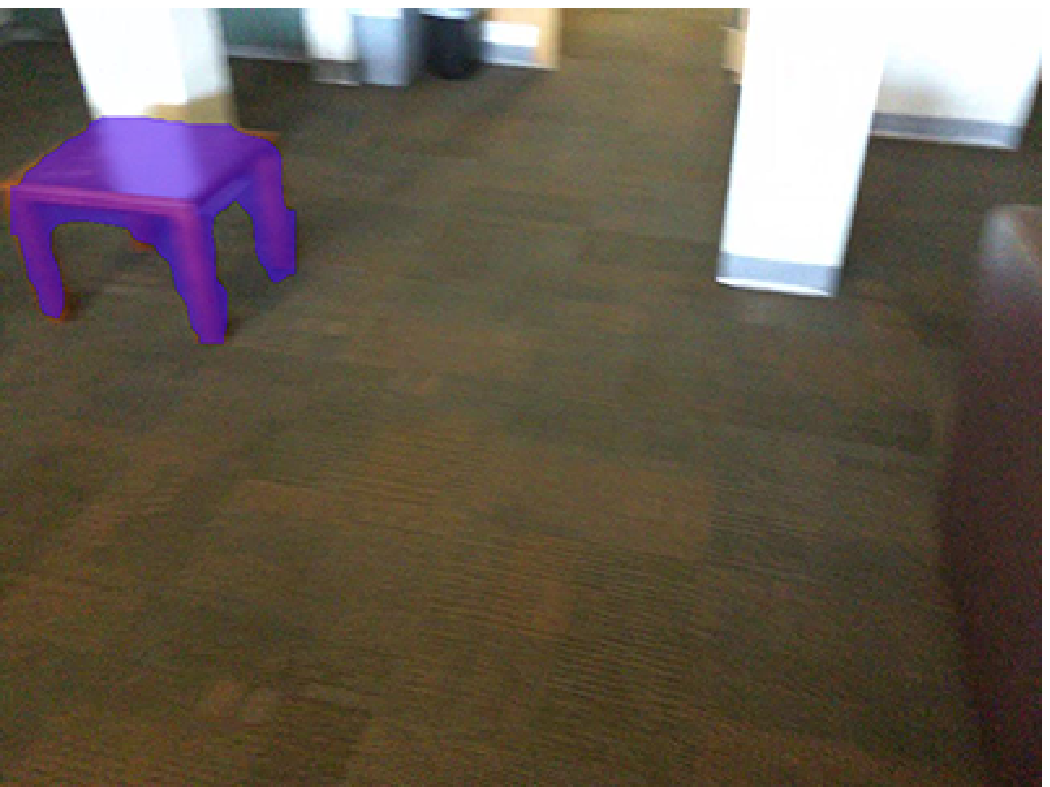}
    \end{minipage}}
  \subfigure[]{
    \centering
    \label{fig:scannet:e}
    \begin{minipage}[b]{0.15\textwidth}
      \centering
      \includegraphics[scale=0.26]{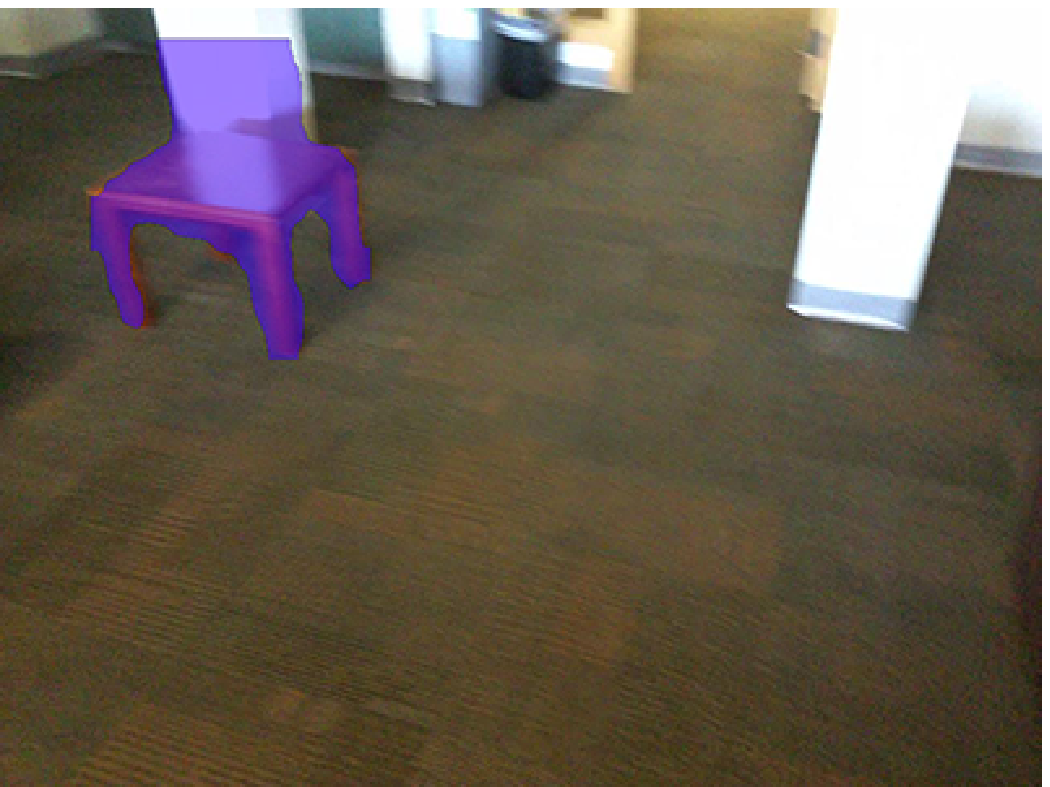}
    \end{minipage}}
  \subfigure[]{
    \centering
    \label{fig:scannet:f}
    \begin{minipage}[b]{0.15\textwidth}
      \centering
      \includegraphics[scale=0.26]{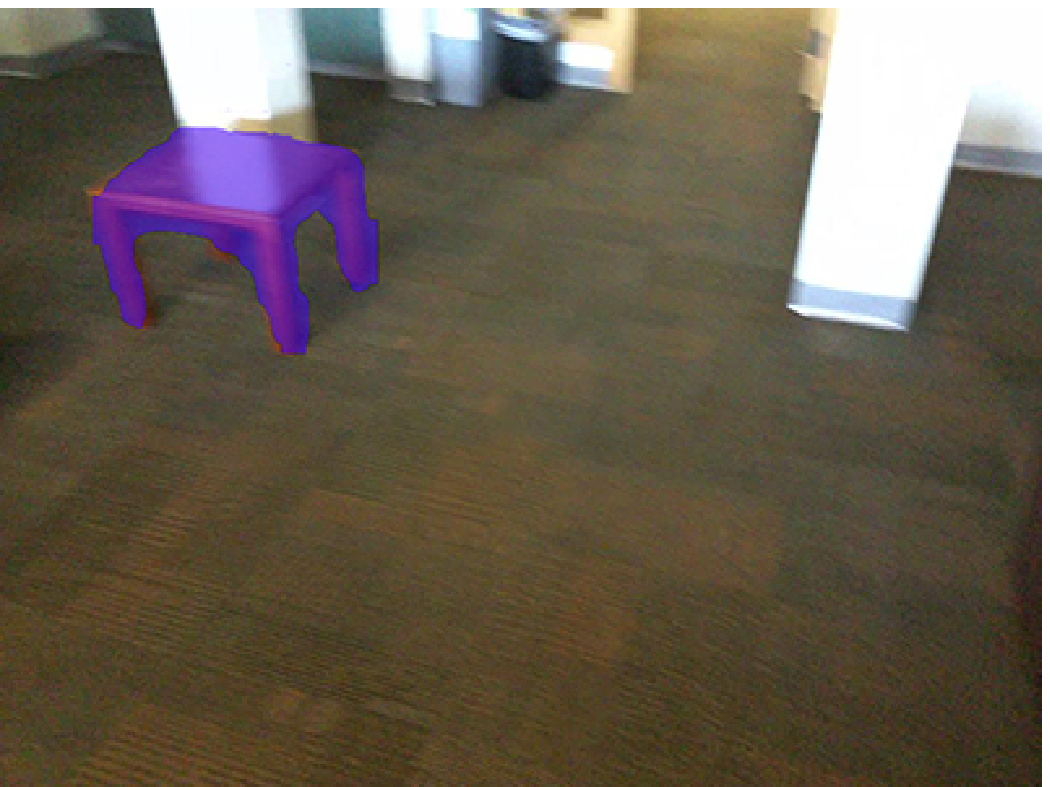}
    \end{minipage}}
  \caption{Examples of the refinement of segmentation.
  (a)-(c): The results of segmentation of last frame, initial segmentation of current frame with a missing part, and refined segmentation of the current frame respectively.
  (e)-(f): The results of segmentation of last frame, initial segmentation of current frame with an oversized part, and refined segmentation of the current frame respectively.
  }
  \label{fig:scannet}
\end{figure}

We selected two example groups of segmented images to illustrate the refinement of segmentation of our algorithm in Fig.~\ref{fig:scannet}. In Fig.~\ref{fig:scannet:b}, a table failed to be segmented probably due to motion blur is added to the refined result (see Fig.~\ref{fig:scannet:c}) by projecting and adding the segmented part of the last frame (see Fig.~\ref{fig:scannet:a}) onto the current one. Fig.~\ref{fig:scannet:f} shows that the oversized segmented area (see the chair in Fig.~\ref{fig:scannet:e}) in the initial segmentation result was shrunk to its correct range by projecting and combining the result in the last frame (see Fig.~\ref{fig:scannet:d}).

The initial segmentation for each frame takes 113 \emph{ms} on average, and the refinement takes about 50 \emph{ms}. The latter process can be further accelerated by utilizing parallel computing or GPU techniques.

\subsection{Test Results on AirSim Generated Dataset}

In the above two tests, the results of our algorithm on performing the two tasks have not been tested simultaneously. Meanwhile, there is also a lack of a test on the relocalization performance of the vSLAM module. Therefore, we created a series of sequences using the Microsoft AirSim simulator~\cite{airsim}. It allows the users to control the movement of a car or UAV in a virtual outdoor environment, and collects the RGBD images as well as other sensor data during the process. The exact pose of the camera and also the exact segmentation results can be generated automatically.

To generate the image sequence data, we select totally 40 different routes in a virtual city area, and run two passes with different camera poses and moveable objects(vehicles, pedestrians, etc.) which may either be moving or static along each route, by controlling a virtual car. The resolutions for the RGB and depth images obtained from the virtual camera bound to the car are set to $640 \times 480$, with frame rate of 15 \emph{fps}. The lengths of routes range from 160 \emph{m} to 400 \emph{m}. There are totally 16 classes in the segmentation results, and they are also mapped to the MS COCO 80 classes.
\begin{table}[htbp]
\caption{Comparison of ORB-SLAM2~\cite{ORBSLAM}  and our vSLAM module on AirSim generated sequences.}
\begin{center}
\begin{tabular}{|c|c|c|c|c|c|c|}
\hline
\multicolumn{1}{|c|}{\multirow{2}{*}{}} &
\multicolumn{3}{|c|}{ORB-SLAM2} & \multicolumn{3}{|c|}{Our vSLAM module}\\
\cline{2-7}
& median & min & max& median & min & max\\
\hline ATE[m]&0.82&0.43&1.03&\textbf{0.39}&0.28&0.61\\
\hline
\end{tabular}
\label{table:relo} 
\end{center}
\end{table}

To test the precision of relocalization in vSLAM, we use the long-term map created in the first pass to compute the fine tracking in the second pass, and compare the ATE[m] of our vSLAM module and that of ORB-SLAM2. It can be seen from the results shown in Table~\ref{table:relo} that our vSLAM module is much better than those of ORB-SLAM2. Note that the ATE[m] values of the tracking results of the AirSim generated dataset is much higher than those of the TUM dataset. This is because the areas of the outdoor scenes in AirSim are much larger than those in TUM which are only limited regions indoors.


\begin{figure} [htbp]
  \centering
  \subfigure[]{
    \centering
    \label{fig:airsimslam:a} 
    \begin{minipage}[b]{0.36\textwidth}
      \centering
      \includegraphics[scale=0.68]{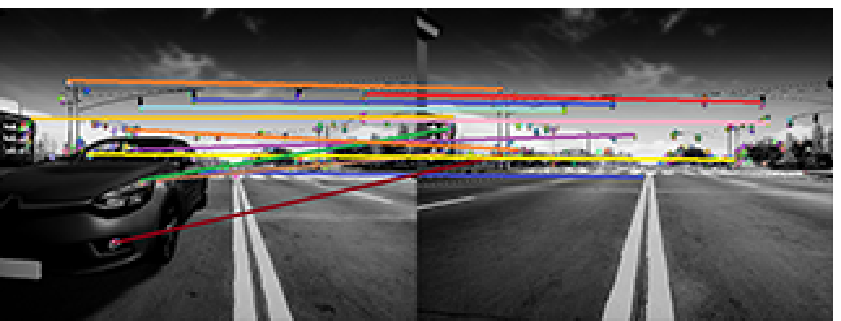}
    \end{minipage}}
    \\
  \subfigure[]{
    \centering
    \label{fig:airsimslam:b}
    \begin{minipage}[b]{0.36\textwidth}
      \centering
      \includegraphics[scale=0.68]{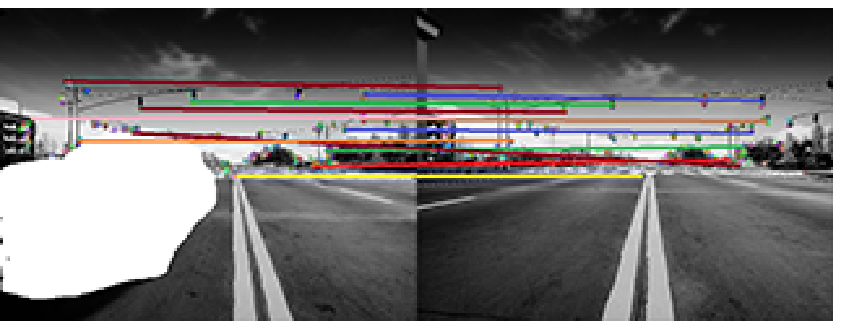}
    \end{minipage}}
  \caption{Feature point matching for two adjacent frames with (a) and without (b) segmenting dynamic objects.}
  \label{fig:airsimslam}
\end{figure}

We show the matching points between two consecutive frames in the generated dataset in Fig.~\ref{fig:airsimslam}. It can be seen that the car contains some feature points which will be mapped to incorrect positions if the car disappears (see Fig.~\ref{fig:airsimslam:a}). By segmenting the car and excluding the feature points (see Fig.~\ref{fig:airsimslam:b}) on it during tracking and mapping, the tracking precision when revisiting the same region will be enhanced.

We also evaluate the performance on segmentation of our algorithm on all the 40 sequences of the second pass, and list the results in Table~\ref{table:segair}. It can be seen that by making use of the pose information to refine the initial segmentation result, our algorithm enhances the accuracy of segmentation.

\begin{table}[htbp]
\caption{Comparison of FCIS~\cite{XXX5} and our segmentation module on AirSim generated dataset.}
\begin{center}
\begin{tabular}{|c|c|c|}
\hline &FCIS&Our segmentation module\\
\hline mAP&0.6702&\textbf{0.6893}\\
\hline mIoU&0.6491&\textbf{0.6611}\\
\hline
\end{tabular}
\label{table:segair} 
\end{center}
\end{table}

From the results tested on three different datasets, it can be seen that our framework effectively improves the precision of vSLAM and segmentation in both indoor and outdoor environment. The performance promotion of the two modules are more obvious for scenes with objects in motion in the current scan or relocated in further scans.



\section{CONCLUSIONS}\label{sec:conc}

We present a unified framework for combining the vision-based localization and segmentation tasks for robotics.  An accurate pose can be refined from the coarse one by identifying and handling the moving and possibly moveable objects respectively with the help of the initial segmentation result, and it further helps to remedy the errors and boundary inaccuracy of the segmented regions to get a more precise segmentation result. Experimental results on various datasets show that our approach is able to make enhancements to both the localization and segmentation for different environments, especially those with dynamic objects and obvious changes. The proposed framework has the potential to be applied to many robotic applications which use vision sensors for synthesized tasks, including autonomous driving, UAV, logistic robots, etc.

\bibliographystyle{IEEEtran}
\bibliography{IEEEabrv,IEEEexample}

\end{document}